\documentclass[11pt]{article}
\usepackage[margin=1.2in]{geometry}
\usepackage{amsmath, amsthm, amssymb}
\usepackage{booktabs}
\usepackage{graphicx}
\usepackage{xcolor}
\usepackage{hyperref}
\usepackage{natbib}
\usepackage{microtype}
\usepackage{parskip}
\usepackage{algorithm}
\usepackage{algorithmic}
\hypersetup{
  colorlinks=true,
  linkcolor=blue!70!black,
  citecolor=blue!70!black,
  urlcolor=blue!70!black
}
\newtheorem{definition}{Definition}

\newtheorem{remark}{Remark}
\newcommand{\eff}{\mathrm{eff}}
\newcommand{\erank}{\rho_{\eff}}
\newcommand{\drift}{\Delta_{\mathrm{CKA}}}

\newcommand{\R}{\mathbb{R}}

\title{\textbf{Rank Collapse, Fixed Points, and the Renormalization Group\\
Structure of MLP Residual Networks}}

\author{
    Parviz Haggi-Mani\textsuperscript{1,2},  Irina Rish\textsuperscript{1, 2}\\
    haggimpa, irina.rish@mila.quebec\\
     \textsuperscript{1}\textnormal{Université de Montréal}, 
    \textsuperscript{2}\textnormal{Mila - Quebec AI Institute}
}
\date{\today}

\begin{document}
\maketitle
\begin{abstract}

The analogy between deep neural network forward passes and
renormalization group theory (RG) flows has been repeatedly noted in the
literature \citep{mehta2014exact, beny2013renormalization,
bordelon2024renormalization, symmetrybroken2023transformer,
alpay2026latentobjectpermanencetopological,
tian2026rgmemrenormalizationgroupinspiredmemory},
but existing treatments remain qualitative and metaphorical: while
depth is described as the equivalent of an RG coarse-graining scale
\citep{wilson1971renormalization, wilson1974renormalization,
symmetrybroken2023transformer}, attention is likened to a partition
function in statistical physics
\citep{symmetrybroken2023transformer,
alpay2026latentobjectpermanencetopological}, and representations are
said to follow an RG flow toward fixed points
\citep{bordelon2024renormalization, wilson1974renormalization}.

In this paper, we move beyond a qualitative description and to an empirical study of the simplest architecture for which the analogy is tractable: a pure MLP residual stack trained on masked token prediction \citep{devlin2019bert} over synthetic Markov chain sequences with known spectral properties.
We report three findings: (i) the effective rank of the residual stream decreases monotonically with depth after training, consistent with the progressive integration of irrelevant degrees of freedom predicted by Wilsonian RG. (ii) and more significantly, this rank collapse is
\emph{selective}: it occurs for chains with short correlation length
$\xi \approx 1$ but is absent for chains with long correlation length
$\xi \approx 7$, measured at the position level to control for
mean-pooling artifacts. The network correctly identifies which degrees of freedom are relevant to the prediction task and preserves them, compressing only the irrelevant ones. This is precisely the content of the relevance
criterion for the RG. (iii) the inter-layer kernel drift is not uniformly distributed across depth but concentrated at one or two specific layer
transitions, with the remainder of the network operating near a fixed
point. This discrete-transition structure challenges a naive uniform-cascade reading of depth-as-scale while remaining consistent with a
fixed-point interpretation.

Together, these findings constitute the first quantitative,
position-level evidence that MLP residual networks implement
a selective coarse-graining procedure whose behavior is governed by
the spectral structure of the input distribution.
We discuss implications for the broader RG program in large language
models and identify the perturbation-decay experiment as the natural
next step toward extracting RG scaling exponents.
\end{abstract}

\section{Introduction}
\label{sec:intro}

RG is one of the most powerful organizational frameworks in
theoretical physics.
Its central idea is that the behavior of complex systems depends on
the scale at which they are observed.
In this scheme, the macroscopic behavior of a large system is
understood through the repeated process of coarse-graining in
real space: systematically integrating out short-range degrees of
freedom leaves an effective large-scale description governed by the
relevant operators that survive at long scales
\citep{wilson1971renormalization, wilson1974renormalization}.
The framework has been applied to statistical mechanics, quantum
field theory, and condensed matter physics, and has recently
attracted attention as a potential lens for deep learning.
\citet{mehta2014exact} established a formal correspondence between
the variational renormalization group and restricted Boltzmann
machines, and \citet{beny2013renormalization} discussed RG in the
context of quantum information, providing early theoretical
grounding for the analogy.

Several authors have since observed that the forward pass of a
Transformer resembles an RG flow: depth plays the role of the RG
scale, each layer performs a kind of coarse-graining by aggregating
token information, and representations evolve from
syntactically-sensitive to semantically-structured as depth
increases.
\citet{symmetrybroken2023transformer} argued that the Transformer
forward pass implements a Kadanoff--Wilson RG flow, defining the
softmax attention matrix as a Boltzmann partition function and the
feedforward sublayers as integrating irrelevant fluctuations in a
local, site-wise fashion reminiscent of real-space RG.
\citet{alpay2026latentobjectpermanencetopological} connected RG
flows to topological phase transitions in Transformer representation
manifolds.
At the level of training dynamics, \citet{bordelon2024renormalization}
applied RG methods to predict non-Gaussian corrections to neural
scaling laws.
At the level of memory and reasoning, \citet{tian2026rgmemrenormalizationgroupinspiredmemory}
proposed a memory evolution mechanism for language agents inspired
by RG coarse-graining, in which less relevant memories are
progressively compressed across reasoning steps.

These treatments share a common limitation: they operate at the
level of qualitative analogy and neither \citet{symmetrybroken2023transformer}
nor \citet{alpay2026latentobjectpermanencetopological} defines or
measures the RG quantities empirically.
No existing work has (i) defined a measurable RG order parameter
for a neural network forward pass, (ii) tested whether that order
parameter behaves as RG theory predicts when the spectral properties
of the input distribution are varied in a controlled way, or (iii)
used the RG framework to make quantitative predictions that are
then empirically verified.

In this paper, we study the simplest architecture for which the RG
analogy is tractable: a pure MLP residual stack with no attention,
trained on masked token prediction over synthetic sequences drawn
from a Markov chain with analytically known spectral properties.
This controlled setting is chosen deliberately in an attempt to
mimic the controlled experimental settings in physics used as a
blueprint for understanding fundamental laws.
Attention-free architectures have been studied from an expressivity
and generalization perspective \citep{gu2023mamba,
shandirasegaran2026theoreticalanalysismambastraining}, but not
through the lens of RG.
Here, the absence of attention is a deliberate design choice: by
removing cross-token mixing from the forward pass, all inter-token
structure in the learned representations must arise from the
training signal alone, making the coarse-graining interpretation
unambiguous.
The Markov chain corpus provides a ground truth: its correlation
length $\xi = -1/\log|\lambda_2|$, where $\lambda_2$ is the second
eigenvalue of the transition matrix $P$, is analytically known and
can be tuned as an experimental parameter.

We measure two quantities as functions of depth across training
checkpoints: (1) the \emph{effective rank} of the residual stream,
a proxy for the number of active degrees of freedom at each layer,
and (2) the \emph{kernel drift}, the layer-to-layer change in the
centered kernel alignment (CKA) similarity matrix, which measures
how much the representational geometry changes at each step.

This paper is organized as follows:
Section~\ref{sec:background} introduces the Markov chain corpus,
the architecture, and the training procedure.
Section~\ref{sec:theory} gives formal definitions of the two
measurement quantities and their connection to RG concepts.
Sections~\ref{sec:rank_collapse}--\ref{sec:discrete_transitions}
present the experimental findings.
Section~\ref{sec:discussion} interprets the results, discusses
limitations, and outlines the perturbation-decay experiment as
the natural next step.

\section{Background}
\label{sec:background}

\subsection{Synthetic corpus: Markov chain sequences}
\label{sec:markov}

Let $P \in \R^{V \times V}$ be a row-stochastic transition matrix
over a vocabulary of $V$ tokens.
We generate sequences $\mathbf{s} = (s_1, \ldots, s_T)$ by sampling
$s_1$ from the stationary distribution $\pi$ (the normalized left
eigenvector of $P$ for eigenvalue 1) and drawing each subsequent
token as $s_{t+1} \sim P_{s_t, \cdot}$.

The spectral gap of $P$ controls the rate at which correlations
decay.
For an ergodic aperiodic chain, correlations at lag $k$ decay as
$C(k) \approx |\lambda_2|^k$, where $\lambda_2$ is the
second-largest eigenvalue by magnitude.
The correlation length is therefore
\begin{equation}
  \xi = \frac{-1}{\log |\lambda_2|}.
  \label{eq:corr_length}
\end{equation}
Intuitively, $\xi$ is the "memory" of the chain; a short $\xi$ means that the chain forgets its past quickly, and a long $\xi$ means the chain has long-range memory i.e., tokens far apart in the sequence are still statistically related. We use two experimental regimes:
\begin{itemize}
  \item \textbf{Short-$\xi$}: $P$ drawn with Dirichlet concentration
        $\alpha = 10$, giving near-uniform rows, $\lambda_2 \approx
        0.44$, and $\xi \approx 1.2$.
        Sequences decorrelate within $\sim$5 steps; the context
        window carries little predictive signal.
  \item \textbf{Long-$\xi$}: $P$ drawn with Dirichlet concentration
        $\alpha = 0.05$, giving sparse rows,
        $\lambda_2 \approx 0.86$, and $\xi \approx 6.7$.
        The mixing time for $\varepsilon = 0.01$ is $\sim$31 steps;
        the context window carries substantial predictive signal.
\end{itemize}

In all experiments, $V = 16$, $T = 64$, and sequences are encoded
as one-hot vectors, giving inputs $X \in \R^{B \times T \times V}$, where $B$ is the batch size.

\subsection{Architecture: MLP residual stack}
\label{sec:arch}

We study a \emph{pre-norm MLP residual stack} with no attention.
The model consists of:
\begin{enumerate}
  \item A projection $\phi_{\mathrm{in}} : \R^V \to \R^d$
        (linear, no bias) mapping one-hot inputs to the model
        dimension $d$.
  \item $L$ residual blocks, each computing
        $x \leftarrow x + \mathrm{MLP}(\mathrm{LayerNorm}(x))$,
        where $\mathrm{MLP}$ is a two-layer network with hidden
        dimension $4d$ and GELU activation.
  \item A final LayerNorm.
  \item A classification head $\phi_{\mathrm{head}} : \R^d \to \R^V$
        (linear) predicting the masked token.
\end{enumerate}
The forward pass returns hidden states
$\{h^{(0)}, h^{(1)}, \ldots, h^{(L)}\}$, where $h^{(0)} =
\phi_{\mathrm{in}}(X)$ and $h^{(l)}$ is the output of block $l$.
All experiments use $d = 64$, $L = 6$, unless otherwise noted.

The absence of attention mechanism is a deliberate design choice.
Without cross-token mixing in the forward pass, any
inter-token structure in the representations must arise from the
training signal alone, not from architectural inductive bias.
This makes the coarse-graining interpretation cleaner: if the model
learns to compress or preserve degrees of freedom, it is because the
loss function demands it, not because the architecture enforces it.

\subsection{Training: masked token prediction}
\label{sec:training}

We train with a BERT-style masked token prediction objective
\citep{devlin2019bert}.
At each step, 15\% of the positions in each sequence are selected
uniformly at random and replaced with the zero vector (which is
unambiguously out-of-distribution from any one-hot input).
The model predicts the original token at masked positions via the
classification head, and the loss is cross-entropy averaged over
masked positions only.

Fresh sequences are sampled from the Markov chain at each training
step, so the model never overfits a fixed dataset.
We train for 10{,}000 steps with batch size 32 and the Adam
optimizer \citep{kingma2017adammethodstochasticoptimization} at learning rate $3 \times
10^{-4}$.

\section{Measurement Framework}
\label{sec:theory}

We define two quantities that serve as our empirical proxies for RG
flow behavior.

\subsection{Effective rank}
\label{sec:erank_def}

Let $H^{(l)} \in \R^{N \times d}$ be the matrix of residual stream
representations at layer $l$, where $N$ is the number of evaluation
tokens (positions $\times$ sequences).
Let $\sigma_1 \geq \sigma_2 \geq \cdots \geq \sigma_d \geq 0$ be
the singular values of $H^{(l)}$.
Define the normalized singular value distribution
$p_i = \sigma_i / \sum_j \sigma_j$.

\begin{definition}[Effective rank]
The effective rank of $H^{(l)}$ is
\begin{equation}
  \erank(H^{(l)}) = \exp\!\left(-\sum_{i} p_i \log p_i\right),
  \label{eq:erank}
\end{equation}
\end{definition}
where we assume that singular values may be close but not exactly zero. 
This definition, due to \citet{roy2007effective}, measures the
entropy of the singular value spectrum, normalized so that
$\erank \in [1, d]$.
A representation with all singular values equal has $\erank = d$
(full rank); a rank-1 representation has $\erank = 1$.

\paragraph{In RG language} $\erank(H^{(l)})$ counts the effective number of
active degrees of freedom at depth $l$.
A decreasing profile $\erank(H^{(0)}) > \erank(H^{(1)}) > \cdots >
\erank(H^{(L)})$ is the signature of progressive elimination of
irrelevant directions i.e., the RG coarse-graining prediction.

\begin{remark}
Note that $H^{(l)}$ is computed using \emph{position-level} (flattened)
representations: each of the $B \times T$ token positions contributes
one row. In section~\ref{sec:selective_compression}, we will demonstrate that measurement modes can conflate the spectral structure of the transition matrix with the diversity of the evaluation batch, giving rise to an artifactual rank collapse.  
\end{remark}

\subsection{Kernel drift}
\label{sec:drift_def}

Let $K^{(l)} \in \R^{N \times N}$ be the centered linear kernel
matrix at layer $l$ (also refer to as gram matrix):
\begin{equation}
  K^{(l)} = H_c^{(l)} {H_c^{(l)}}^\top,
  \quad
  H_c^{(l)} = (I - \tfrac{1}{N}\mathbf{1}\mathbf{1}^\top) H^{(l)}
\end{equation}
where $\mathbf{1}$ is a column vector of all ones, and the columns of $H_c^{(l)}$ average to 0.
\begin{definition}[Kernel drift]
The kernel drift between consecutive layers $l$ and $l+1$ is
\begin{equation}
  \drift(l, l+1) = 1 - \mathrm{CKA}(H^{(l)}, H^{(l+1)}),
  \label{eq:drift}
\end{equation}
where 
\begin{equation}
    \mathrm{CKA}(A, B) = \langle K^A, K^B \rangle_F /
(\|K^A\|_F \|K^B\|_F)
\end{equation} is the linear centered kernel alignment
\citep{kornblith2019similarity}, 
and the Frobenius inner product, defined as $\langle K^A, K^B \rangle_F = \sum_{i,j} K^A_{ij} K^B_{ij}$,
measures how similar two $N \times N$
kernel matrices are entry-by-entry.
\end{definition}

$\drift(l, l+1) = 0$ if and only if the representational geometry
is identical at consecutive layers; $\drift(l, l+1) = 1$ if the
representations are geometrically orthogonal.
A small drift indicates the layer is near a fixed point of the flow.

In RG language, the drift profile $\{\drift(l, l+1)\}_{l=0}^{L-1}$
describes where in the network active representational change is
occurring.
A uniform, decreasing profile would be consistent with a smooth RG
flow. Concentration of drift at isolated transitions would indicate a
discontinuous or phase-transition-like reorganization.

\section{Experiment 1: Rank Collapse for Short-\texorpdfstring{$\xi$}{\text xi} Chains}
\label{sec:rank_collapse}

We train the MLP residual stack on the short-$\xi$ corpus
($\xi \approx 1.2$) and measure the effective rank profile at
11 checkpoints evenly spaced over 10{,}000 training steps.
Representations are extracted on a fixed evaluation set of 64
sequences (4{,}096 token positions) with no masking applied.

Table~\ref{tab:rank_short} reports the effective rank at each layer
at initialization and at the final checkpoint. At initialization, the effective rank profile is flat with a compression ratio $\erank(H^{(0)}) / \erank(H^{(L)})$ of $1.05\times$, as expected
for randomly initialized weights.
After training, the profile exhibits a clear monotone decrease at every layer,
with an $8.4\times$ compression ratio and a collapse at layer 3.
By the final layer the representation is approximately $1D$,
consistent with convergence to a low-dimensional fixed-point
representation encoding the stationary distribution.

\begin{figure}[ht]
  \centering
  \begin{minipage}[t]{0.48\linewidth}
    \centering
    \includegraphics[width=\linewidth]{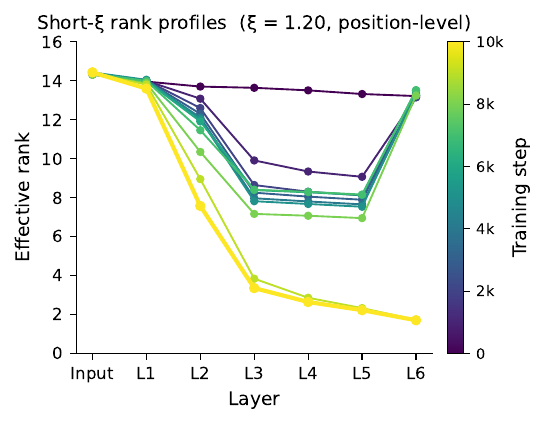}
    \caption{Effective rank profiles across depth for the short-$\xi$
    chain ($\xi = 1.2$, flatten mode), at 11 training checkpoints 
    from initialization (dark) to step 10{,}000 (light).
    Rank is near-uniform at initialization and collapses monotonically
    with depth after training, with a sharp fan visible between
    steps 8{,}000--9{,}000 at layers $L3$--$L6$.}
    \label{fig:rank_short}
  \end{minipage}
  \hfill
  \begin{minipage}[t]{0.48\linewidth}
    \centering
    \includegraphics[width=\linewidth]{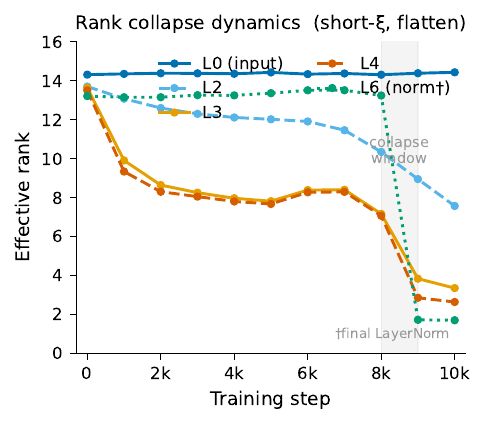}
    \caption{Effective rank at selected layers as a function of
    training step, short-$\xi$ chain (flatten mode).
    $L0$ (control) remains flat throughout.
    $L3$ and $L4$ are stable until the shaded collapse window
    (steps 8{,}000--9{,}000), where they drop abruptly.
    The sharpness of the transition is inconsistent with a smooth
    RG cascade and suggests a phase-transition-like reorganisation
    during late training.}
    \label{fig:collapse_transition}
  \end{minipage}
\end{figure}

Tracking the rank profile at intermediate checkpoints reveals that
the collapse is not gradual: the rank at $L3$ remains near 8
through step 8{,}000, then drops sharply to 3.4 between steps
9{,}000 and 10{,}000 (see Figure \ref{fig:rank_short} and the collapse window Figure \ref{fig:collapse_transition}).
This late, sharp event seems inconsistent with a smooth RG cascade, and suggests a phase-transition-like reorganization during late training.

\begin{table}[ht]
\centering
\caption{Effective rank across depth, short-$\xi$ chain ($\xi=1.2$),
flatten mode}
\label{tab:rank_short}
\begin{tabular}{lccccccc|cc}
\toprule
 & $L0$ & $L1$ & $L2$ & $L3$ & $L4$ & $L5$ & $L6$
 & ratio & collapse \\
\midrule
Init  & 16.0 & 15.8 & 15.7 & 15.5 & 15.4 & 15.3 & 15.2
      & $1.05\times$ & --- \\
Final & 14.4 & 13.6 & 7.6  & 3.4  & 2.6  & 2.2  & 1.7
      & $8.4\times$ & $L3$ \\
\bottomrule
\end{tabular}
\end{table}

\section{Experiment 2: Selective Compression for Long-\texorpdfstring{$\xi$}{\text xi} Chains}
\label{sec:selective_compression}

Running the same experiment for chains with a long correlation length $\xi\approx 6.7$, we note that the results depend on how effective ran is measured. First, let's note that a long/short correlation length means that the chain remembers its past for many/few steps. This is equivalent to saying that the chain is slow/fast-mixing. The mixing time is formally defined as the time it takes for the chain to approach its stationary distribution $\pi$ from an arbitrary starting point:
\begin{equation}
t_{\mathrm{mix}}(\varepsilon) = \xi \cdot \log(1/\varepsilon),
\end{equation}
where $\varepsilon$ is a hyperparameter that determines the chosen distance to the stationary distribution.  For example, the long-$\xi\approx 6.7$ chain with the hyperparameter $\varepsilon=0.01$ is equivalent to a mixing time of $\sim$31 steps.  

\subsection{Measuring the Effective Rank Correctly}
Effective rank can be measured in two ways, either at position-level (flatten mode here) or after mean-pooling (mean over the sequence length). In the latter case, instead of keeping all 4096 individual token positions as separate rows, mean-pool collapses each sequence of 64 tokens into a single vector by averaging. So the usual $4096 \times 64$ matrix in flatten mode, collapses to a $64 \times 64$ matrix, one row per sequence. Table~\ref{tab:rank_both} reports the results for both chains. 

\begin{table}[ht]
\centering
\caption{Effective rank at final checkpoint under two measurement
modes.
Pool = mean over sequence length; Flat = position-level (flatten).}
\label{tab:rank_both}
\begin{tabular}{llccccccc|cc}
\toprule
Run & Mode & $L0$ & $L1$ & $L2$ & $L3$ & $L4$ & $L5$ & $L6$
    & ratio & collapse \\
\midrule
Short-$\xi$ & pool & 13.6 & 12.1 & 3.3  & 2.2 & 1.6 & 1.5 & 1.3
            & $10.6\times$ & $L2$ \\
Long-$\xi$  & pool &  7.4 &  6.8 & 5.8  & 1.9 & 1.4 & 1.2 & 1.1
            & $6.5\times$  & $L3$ \\
\midrule
Short-$\xi$ & flat & 14.4 & 13.6 & 7.6  & 3.4 & 2.6 & 2.2 & 1.7
            & $8.4\times$  & $L3$ \\
Long-$\xi$  & flat & 11.2 & 10.8 & 8.1  & 8.1 & 8.1 & 8.0 & 10.8
            & $1.0\times$  & none \\
\bottomrule
\end{tabular}
\end{table}

The two-by-two comparison reveals that this poses a problem for the slow-mixing i.e., long-$\xi$ chains. They tend to stay in certain regions of the state space for a long time before moving on (roughly 31 steps here). Because of this, the 64 mean-pooled vectors naturally cluster into a small number of groups, and the gram matrix of these 64 vectors therefore looks low-rank (roughly equal to the number of dominant states). However, this is not because the network compressed the representations but because the sequences themselves naturally clustered. In conclusion, the $6.5\times$ compression in pool mode is entirely an artifact of the slow-mixing sequence clustering by dominant states.
\begin{figure}[ht]
  \centering
  \includegraphics[width=.90\linewidth]{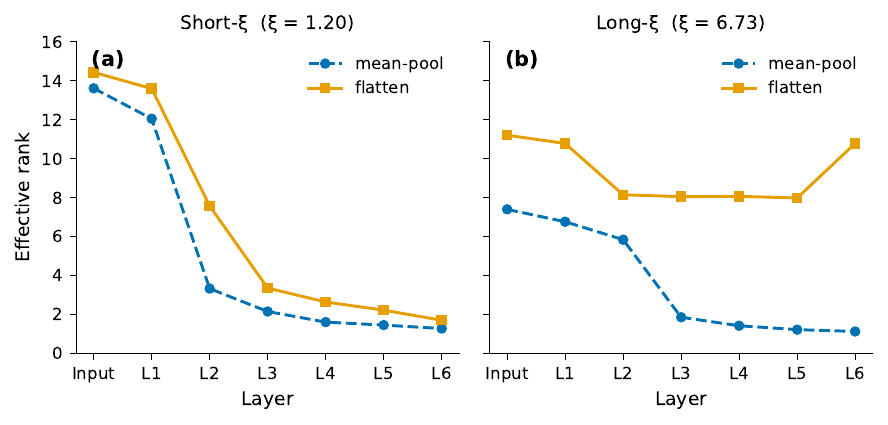}
  \caption{Effective rank at the final checkpoint under pool (dashed)
  and flatten (solid) measurement modes.
  \textbf{(a)} Short-$\xi$: the two modes nearly coincide; the
  confound is small.
  \textbf{(b)} Long-$\xi$: pool mode shows apparent collapse to
  rank~$\approx 1$; flatten mode shows no collapse (rank~$\approx
  8$--$11$).
  The discrepancy in panel~(b) is a mean-pooling artifact arising
  from the low token diversity of slow-mixing sequences.}
  \label{fig:pool_vs_flat}
\end{figure}

Note that mean-pooling does not cause an equally big problem for the fast-mixing i.e., short-$\xi$ chains. In this case, since mixing occurs in about 5 steps within any sequence of 64 tokens, the chain visits all states many times and the average vector ends up looking similar for every sequence (roughly the stationary distribution $\pi$). The mean-pooled vectors do not cluster in the same way for fast mixing chains, and the artifact is much smaller.

The flatten mode measurement removes this by keeping every token position separate, so the gram matrix reflects the actual geometry of the representations for both types of chain rather than the statistical structure of the sequences. As seen above, the long-$\xi$ chains in flatten mode show a compression ratio of $1.0\times$, i.e., the network is not compressing at all. However, we observe rank collapse for the short-$\xi$ chains. 

\begin{figure}[ht]
  \centering
  \includegraphics[width=0.48\linewidth]{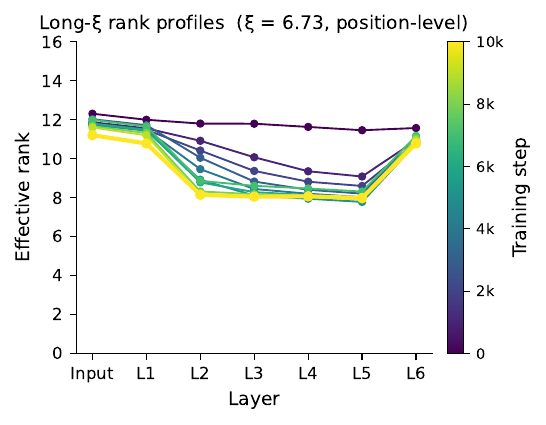}
  \caption{Effective rank profiles across depth for the long-$\xi$ chain
  ($\xi \approx 6.7$, flatten mode) at 10 training checkpoints.
  Unlike the short-$\xi$ chain(Figure~\ref{fig:rank_short}), the effective rank remains stable across
  all layers, confirming that no rank collapse occurs when the input
  distribution contains long-range correlations.}
  \label{fig:rank_long}
\end{figure}

Clearly, if we had only used mean-pooling, we would have found a diametrically different and contradictory result, as both chains would have shown a strong rank collapse (compression ratio $6.5\times$) and an apparent collapse at $L3$, both of which have indicated a strong RG compression. This conclusion would have been completely wrong. The correct conclusion, visible only in flatten mode, is that the short-$\xi$ chains compress strongly and the long-$\xi$
chains do not compress at all (Figure \ref{fig:rank_long}). This differential behavior is the central finding of the paper.

\paragraph{In RG language:}the degrees of freedom that survive to the final
layer are exactly those that are relevant to the prediction task.
For a short-$\xi$ chain, nearly all positional degrees of freedom
decorrelate within $\sim$5 steps and are irrelevant to predicting
any masked token; the model discards them, and rank collapses.
For a long-$\xi$ chain, positional information is relevant across
the full context window; the model preserves it, and rank does
not collapse. Rank collapse is therefore a signature of \emph{irrelevance}, not of learning, and the network implements the RG relevance criterion
correctly.

\subsection{Two Artifacts and One Genuine Signal}
\label{{sec:architecture_artifact}}
Here we discuss two other artifacts related to the current setup. First, since the architecture contains no attention, each token position is processed
independently. A masked token receives the zero vector as input, which clearly does not carry positional or contextual information. Therefore, the theoretically optimal prediction for the network is always $\pi$, the stationary distribution, i.e., the best any position-wise
model with no access to context can achieve. This is the \emph{first artifact}: the rank profiles cannot be measuring context exploitation, because the architecture makes context exploitation impossible.

Furthermore, the two types of chains differ in the sharpness of $\pi$ due to the Dirichlet concentration parameter used to generate $P$. Concentration $\alpha = 10$ produces near-uniform rows and a near-uniform stationary distribution. Concentration $\alpha = 0.05$ produces sparse rows, concentrating $\pi$ around a small number of dominant states. For a zero input $x = \mathbf{0}$, the classification head reduces to
\begin{equation}
  \text{output}_i
  = \frac{e^{(Wx+b)_i}}{\sum_j e^{(Wx+b)_j}}
  = \frac{e^{b_i}}{\sum_j e^{b_j}}
  = \mathrm{softmax}(b)_i,
\end{equation}
since $W \cdot \mathbf{0} = 0$. This makes the weight matrix obsolete, so the entire prediction is determined by the bias vector $b$ so that it satisfies  $\mathrm{softmax}(b) = \pi$.

\begin{figure}[ht]
  \centering
  \includegraphics[width=0.85\linewidth]{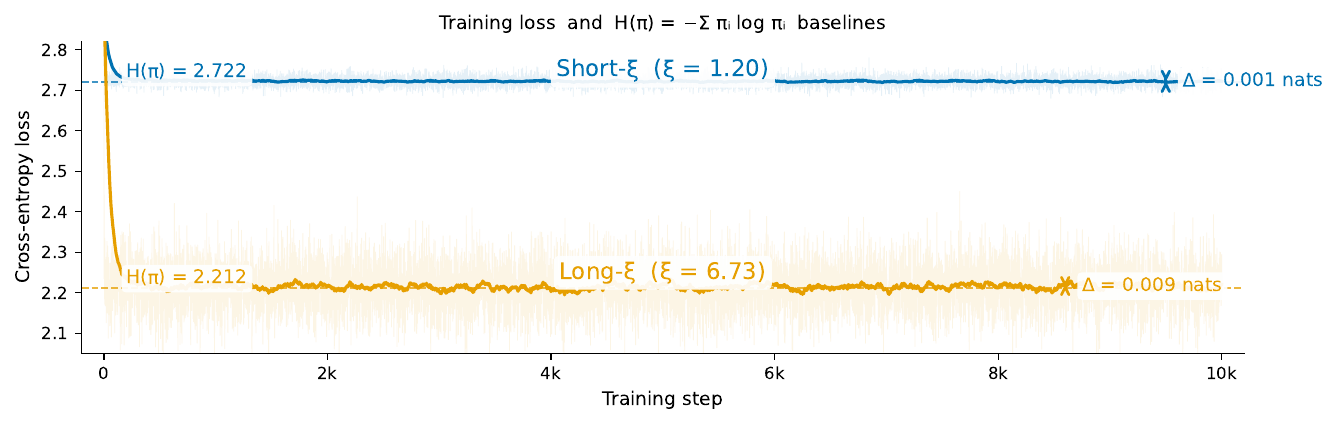}
  \caption{Training loss curves for the short-$\xi$ (blue, $\xi = 1.20$) and
  long-$\xi$ (orange, $\xi = 6.73$) chains over 10{,}000 steps. Faint texture
  shows raw per-step loss; solid lines show exponential moving average (EMA-smoothed) loss. Dashed horizontal
  lines (left) mark $H(\pi) = -\sum_i \pi_i \log \pi_i$ for each chain ($H(\pi) = 2.722$
  for short-$\xi$; $H(\pi) = 2.212$ for long-$\xi$) --- the lowest loss achievable
  by a position-wise model with no access to context. Double-headed arrows show the
  gap $\Delta$ between the final smoothed loss and the $H(\pi)$ floor. The
  short-$\xi$ model converges to within 0.001 nats of $H(\pi)$; the long-$\xi$
  model ends 0.009 nats above $H(\pi)$, reflecting the greater distance of its
  peaked target distribution from the near-zero initialization of $b$, not a
  failure of training.}
  \label{fig:loss_curves}
\end{figure}

For short-$\xi$, $\pi$ is near-uniform, so the optimal $b$ is near-zero, 
essentially where random initialization already places it. The network converges in very few steps, leaving a gap of only $\Delta = 0.001$ nats above $H(\pi)$. For long-$\xi$, $\pi$ is peaked, so $b$ must develop strong preferences across entries, requiring many more gradient steps to move away from initialization. After 10{,}000 steps the gap is $\Delta = 0.009$ nats, not because the data is harder to learn in any general sense, but because the target distribution is further from where the bias starts. This is the \emph{second artifact}: the larger loss gap for the long-$\xi$ chain is an initialization artifact, not evidence of a learning difference (Figure \ref{fig:loss_curves}).

With both artifacts identified and set aside, the rank profiles carry a genuine
signal. They measure how the residual stack organizes its representations as a
function of the input distribution's statistical structure, when learning to output
a fixed target from a zero input. The short-$\xi$ chain produces near-exchangeable
sequences: the model converges to a near-identical output at every position, and
positional representations collapse to a single direction ($\erank \approx 1.7$ at
the final layer). The long-$\xi$ chain produces sequences with persistent local
structure: representations at different positions remain geometrically distinct
throughout the stack ($\erank \approx 8$--$11$ across all layers), even though the
model cannot exploit this structure for prediction.

\paragraph{In RG language:} the short-$\xi$ sequences are dominated by irrelevant
degrees of freedom; they decorrelate within $\sim$5 steps, so most positional
dimensions carry no persistent structure and are integrated out. The long-$\xi$
sequences retain positional structure across the full context window; the
representations encode this structure geometrically, even without a mechanism to use it, and rank is preserved. The compression behavior is therefore determined by the correlation length of the input distribution, not by the architecture, and not by the loss gap.

\section{Experiment 3: Discrete Transition Structure}
\label{sec:discrete_transitions}
In this experiment, we focus on the structure of the transitions for the long- and short-$\xi$ chains. Table~\ref{tab:drift} reports the kernel drift profile at the final checkpoint for all four experimental conditions.

\begin{table}[ht]
\centering
\caption{Kernel drift $\drift(l, l+1)$ at final checkpoint.
Mono = fraction of consecutive drift pairs that are decreasing.}
\label{tab:drift}
\begin{tabular}{llcccccc|c}
\toprule
Run & Mode & $0{\to}1$ & $1{\to}2$ & $2{\to}3$ & $3{\to}4$
    & $4{\to}5$ & $5{\to}6$ & mono \\
\midrule
Short-$\xi$ & pool
  & 0.276 & 0.107 & 0.083 & 0.014 & 0.001 & 0.009 & 0.80 \\
Long-$\xi$  & pool
  & 0.005 & 0.090 & 0.008 & 0.005 & 0.005 & 0.003 & 0.80 \\
\midrule
Short-$\xi$ & flat
  & 0.253 & 0.266 & 0.292 & 0.034 & 0.011 & 0.077 & 0.40 \\
Long-$\xi$  & flat
  & 0.074 & 0.560 & 0.004 & 0.003 & 0.001 & 0.383 & 0.60 \\
\bottomrule
\end{tabular}
\end{table}

\begin{figure}[ht]
  \centering
  \begin{minipage}[t]{0.45\linewidth}
    \centering
    \includegraphics[width=\linewidth]{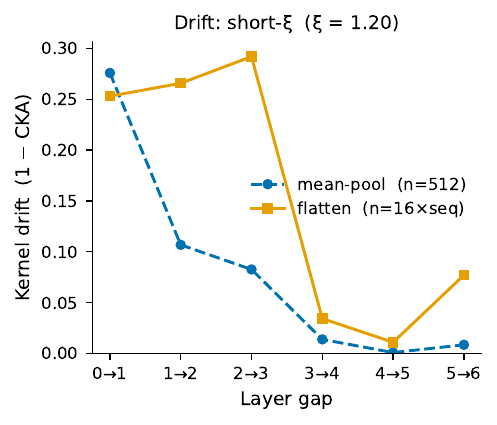}
    \caption{Kernel drift profiles at the final checkpoint under pool
    (dashed) and flatten (solid) modes, short-$\xi$ chain.
    Both modes show a roughly decreasing profile; pool is smoother.}
    \label{fig:drift_short}
  \end{minipage}
  \hfill
  \begin{minipage}[t]{0.45\linewidth}
    \centering
    \includegraphics[width=\linewidth]{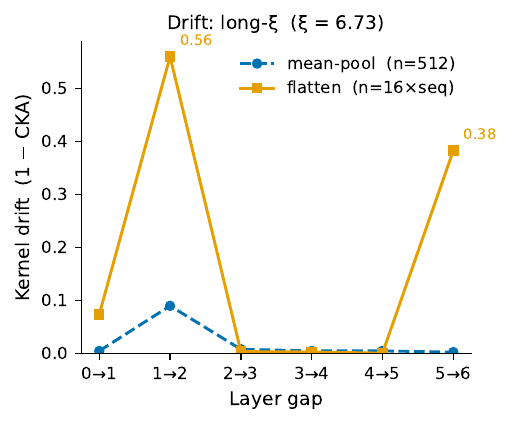}
    \caption{Kernel drift profiles at the final checkpoint under pool
    (dashed) and flatten (solid) modes, long-$\xi$ chain.
    Flatten reveals two isolated spikes at gaps $1{\to}2$
    ($\drift = 0.56$) and $5{\to}6$ ($\drift = 0.38$) with
    near-zero drift in between, exposing the discrete-transition
    structure obscured by mean-pooling.}
    \label{fig:drift_long}
  \end{minipage}
  \caption{Kernel drift profiles at the final checkpoint under pool and flatten modes.}
  \label{fig:drift_profiles}
\end{figure}

\paragraph{Long-$\xi$:}The final kernel drift profile  of the long-$\xi$ chain (Figure \ref{fig:drift_long})(flatten mode, as it does not produce a false rank collapse)
\begin{equation}
0.074 \to 0.560 \to 0.004 \to 0.003 \to 0.001 \to 0.383.
\end{equation}
shows that layers 2 through 5 have near zero kernel drift i.e., they do not contribute to a geometric change of the representations, and that nearly all change in the representations is
concentrated in two transitions: $L1{\to}L2$ (drift $= 0.56$) and
$L5{\to}L6$ (drift $= 0.38$). In other words, the network is not a smooth cascade but a pair of sharp transformations with four nearly-identity layers in between.

As mentioned previously, this is inconsistent with the expectation of a uniform RG flow in the phase space of continuum RG: A naive reading of depth as RG-scale predicts that each layer performs roughly equal coarse-graining i.e., kernel drift should decrease smoothly and monotonically from input to output, with every layer contributing a share of the total transformation. 
However, Table~\ref{tab:drift} and
Figure~\ref{fig:drift_profiles} paint a more complex picture. 
We should pay attention to two issues here: First, one of the features of the phase space in continuum RG theory is fixed points. These are configurations of the system that are invariant under the RG transformation; applying further coarse-graining steps leaves them unchanged. Second, assuming that we are examining a discrete-network analogue of a continuous system, the smooth flows in a continuum RG may not be directly observable. The expectation of a smooth cascade would ideally be realized at the limit of many small steps i.e., an infinitely deep network, which our six layer network is not.  

Therefore, the current observations seem consistent with the notion of fixed points: In this discrete setting, the flow occurs in one step ($L1{\to}L2$). The network simply reaches a fixed point, remains there for four layers ($L2{\to}L5$) (a \emph{fixed-point plateau}), and exits the plateau in a single readout step ($L5{\to}L6$). In other words, with only 6 layers and a simple task, this network solves the problem in the minimum number of transformations and coasts the rest of the way. The near-zero drift across layers ($L2{\to}L5$) says exactly this: the representations have reached a configuration (finding the fixed point) that the MLP blocks leave essentially unchanged until $L5$. 

Note that the drift spike at the ($L5{\to}L6$) transition reflects a difference in what each block is trained to do. The MLP blocks in the plateau do not face direct pressure from the classification head to reorganize the representations, as they already encode useful information, and the loss is minimized by leaving the block unchanged (they act as approximate identity maps). By contrast, $L6$  must bridge the plateau representation and the linear classification head: its weights are trained specifically to make the token embeddings linearly separable, which requires a genuine geometric transformation.

\begin{figure}[ht]
  \centering
  \includegraphics[width=.80\linewidth]{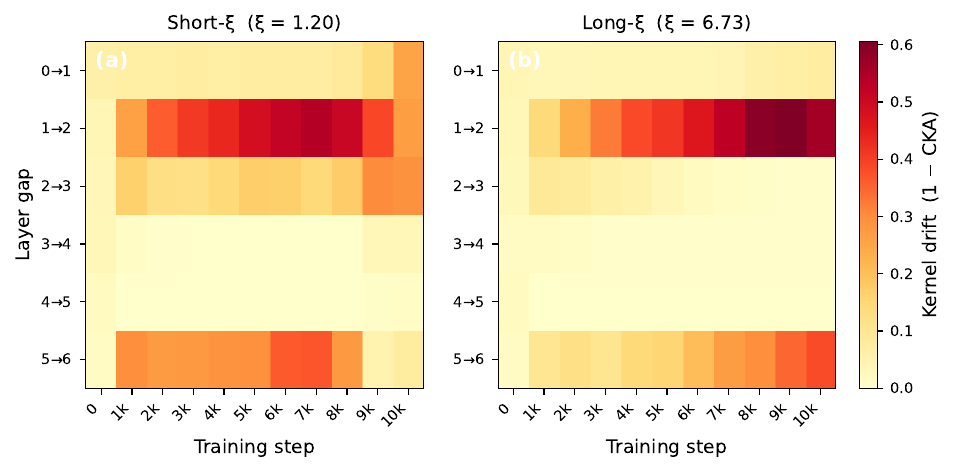}
  \caption{Kernel drift heatmaps (flatten mode) across training.
  Rows = layer gaps; columns = training step; colour = drift magnitude (YlOrRd,
  shared scale 0--0.61); 
    \textbf{(a)} Short-$\xi$: high drift concentrated stably in the
layer gaps ($L1\to L2$, $L2\to L3$) throughout
training, with near-zero drift in gaps $L3{\to}L4$ and $L4{\to}L5$
from the beginning. The active compression zone is fixed from early in training and does not migrate; the fixed-point plateau occupies layers $L3 \to L5$.
  \textbf{(b)} Long-$\xi$: two persistent high-drift rows at gaps
  $L1{\to}L2$ and $L5{\to}L6$, with near-zero drift (pale yellow) in between, stable across all training steps.
  The long-$\xi$ network organizes into a fixed-point plateau $L2\to L5$ with active transformations at its boundaries.}
  \label{fig:drift_heatmap}
\end{figure}

\paragraph{In RG language:}the two drift spikes mark the boundaries of the \emph{fixed-point plateau} with $L1{\to}L2$ being the \emph{entry transition}, and $L5{\to}L6$ as the \emph{exit transition} (a final readout transformation that converts the fixed-point representations into a form suitable for the classification head). Between these two transitions, the network is in principle at rest. Here, the drift heatmap (Figure~\ref{fig:drift_heatmap}) shows two bright rows at
$L1{\to}L2$ and $L5{\to}L6$. These layers remain stable across all training steps: the entry and exit transitions lock in early and do not move. In RG language, this fixed-point plateau is a stable attractor of the training dynamics, and not a transient artifact.

\paragraph{Short-$\xi$:}
The final kernel drift profile of the short-$\xi$ chain in flatten mode
\begin{equation}
0.253 \to 0.266 \to 0.292 \to 0.034 \to 0.011 \to 0.077
\end{equation}
tells a complementary story:
Towards the end of training, high drift is concentrated in layer gaps $L0{\to}L1$, 
$L1{\to}L2$, $L2{\to}L3$, while gaps $L3{\to}L4$ and
$L4{\to}L5$ are near zero throughout training, with a small increase at
$L5{\to}L6$ for the reasons we explained above. The drift heatmap shows a \emph{stable} high drift structure, particularly in layers $L1{\to}L2$ and $L2{\to}L3$ from the beginning of training; the middle gaps $L3{\to}L4$ and $L4{\to}L5$ remain pale throughout.

\paragraph{In RG language:} the short-$\xi$ profile shares the same
\emph{bracketed} structure as the long-$\xi$ case: a multi-step entry
region $L0{\to}L3$, a fixed-point plateau $L3$--$L5$,
and an exit transition $L5{\to}L6$.
The difference is that the plateau is less clean: the small but
non-negligible drift at $L2{\to}L3$ indicates that a residual
amount of compression work spills into the first plateau layer
before the representations fully settle.
This is consistent with the short-$\xi$ chain that presents more
irrelevant variance at the input; the chain decorrelates within
$\sim$5 steps, so a fraction of the irrelevant short-range modes
are not fully integrated out at the entry transition alone and
require one additional step.
Once the plateau is reached at $L3$, however, the behavior is
the same as in the long-$\xi$ case: the MLP blocks leave the
representations essentially unchanged until the exit transition
at $L5{\to}L6$.

\paragraph{A final comparison} shows that the two regimes differ in how much irrelevant information must be discarded before the fixed point is reached.
In the short-$\xi$ chain, the transition matrix rows are nearly
uniform (the Dirichlet concentration parameter $\alpha = 10$ produces rows that are nearly uniform and so each token has roughly equal probability of transitioning to any other token), so most of the variance in the input is short-range noise with no predictive value; the entry transition and the first plateau layer together integrate it out.

In the long-$\xi$ chain, the rows are sparse ($\alpha = 0.05$)
and the dominant token at each position carries genuine predictive
signal across the full context window; almost nothing is irrelevant,
so the entry transition restructures the representation without
discarding it, and the plateau preserves the full dimensionality.
The effective rank results are the direct signature of this
difference: rank collapses by $8.4\times$ in the short-$\xi$ case
and not at all ($1.0\times$) in the long-$\xi$ case.

\paragraph{In conclusion,} these results suggest that the MLP residual stack does not implement an RG flow, but rather its \emph{outcome}: a sparse, efficient solution in which most layers sit at a fixed point and only a minimal number of active transformations are performed. This is the network analogue of the RG fixed point, which is the physically meaningful object, with the flow serving only as the path to reach it. The short-$\xi$
and long-$\xi$ networks differ not in their gross architecture (both converge to the same bracketed plateau structure) but in how cleanly they enter it. Long-range correlations are concentrated enough to be extracted in a single sharp compression step; short-range noise is more diffuse and requires a slightly more distributed entry before the plateau is reached. In both cases, once the plateau begins, the network idles, until the final layer.

\section{Discussion}
\label{sec:discussion}

\paragraph{Three properties confirmed:}
The trained MLP residual networks have three properties consistent with RG theory: (i) The effective rank of the residual stream decreases monotonically with depth for the short-$\xi$ chain ($8.4\times$ collapse from input to final layer), consistent with progressive integration of irrelevant degrees of freedom. (ii) This collapse is absent for the long-$\xi$ chain at the position level ($1.0\times$): when the input contains long-range correlations relevant to the prediction task, the network preserves representational dimensionality rather than discarding it. (iii) Inter-layer kernel drift is concentrated at one or two specific transitions, with the remainder of the network operating near a fixed point. Both chains converge to the same bracketed plateau structure, an entry transition, a near-identity middle, and an exit transition, differing only in how much irrelevant variance is integrated out before the plateau is reached.

These findings constitute quantitative, position-level evidence that a trained MLP residual stack implements the result of a coarse-graining procedure governed by the spectral structure of the input distribution. The network identifies which degrees of freedom are relevant to the prediction task and discards only the irrelevant ones: rank collapse is a signature of irrelevance, not of learning per se. Crucially, this conclusion follows from the training signal alone: the MLP blocks have no built-in inductive bias toward coarse-graining, suggesting that the RG fixed-point structure is a property of the task and the data, not of the architecture, and should persist in deeper and more expressive models.

\paragraph{Three confounds dismissed:}
Alternative explanations for the observed rank profiles were identified and ruled out (Section~\ref{sec:selective_compression}): (i) The apparent rank collapse of the long-$\xi$ chain under mean-pool measurement is a pooling artifact arising from state-clustering in slow-mixing sequences. (ii) The independence of token positions is a consequence of the attention-free architecture, not a driver of compression. The larger loss gap for the long-$\xi$ chain is an initialization artifact, not evidence of a learning difference .The rank profiles therefore constitute a genuine signal about the spectral structure of the data distribution.

\paragraph{Limitations:}
Several limitations apply to the current work: 

(i) the experiments use 6-layer networks with $d=64$, which are
too small to exhibit the full range of RG behavior.
A sweep over $L \in \{6, 12, 24\}$ and $\xi \in \{1, 3, 10, 30\}$
is needed to test whether the collapse depth scales linearly with
$\xi$ as the RG picture predicts (The primary goal of this work was to establish a baseline).  

(ii) there is a 29\% input rank gap between the two chains
under flatten mode (11.2 vs.\ 14.4). This is a genuine confound, arising
from the lower token diversity of slow-mixing sequences.
A fully controlled comparison would require either fixing the
evaluation distribution or injecting perturbations along individual
eigenvectors of $P$ and measuring their layer-wise decay rates. This would simultaneously address the confound and provide
quantitative RG scaling exponents.

(iii) since the model was trained with $15\%$ masked token positions, measuring the representations with no masking applied during evaluation might be slightly out of distribution, and influence the rank and drift measurements.

\section{Conclusion}
\label{sec:conclusion}

We have studied MLP residual stacks trained on masked prediction
over Markov chain sequences, using effective rank and kernel drift
as empirical proxies for RG order parameters.
Three findings emerge: 

First, rank collapse is selective: it occurs when the input
distribution contains short-range correlations ($\xi \approx 1.2$,
collapse ratio $8.4\times$) and is absent when the distribution
contains long-range correlations ($\xi \approx 6.7$, collapse ratio
$1.0\times$), measured at the position level to control for
mean-pooling artifacts. 
The network discards only the degrees of freedom that are irrelevant
to the prediction task and preserves those that are relevant;
precisely the content of the Wilsonian RG relevance criterion.

Second, the inter-layer kernel drift is not distributed uniformly
across depth but concentrated at one or two specific transitions,
with the remainder of the network operating near a fixed point.
Both the short-$\xi$ and long-$\xi$ networks converge to a
bracketed fixed-point plateau structure: an entry transition, a
near-identity middle, and an exit transition.
The two regimes differ only in the cleanliness of the entry: the
long-$\xi$ network reaches its fixed point in a single sharp step,
while the short-$\xi$ network requires one additional layer to
complete the compression of irrelevant short-range modes.

Third, and most importantly, the fixed-point plateau is a stable
attractor of the training dynamics.
The boundaries between the active-transformation layers and the
plateau lock in early and do not shift as training continues.
This suggests that the sparse fixed-point structure is determined
by the spectral properties of the data-generating process, not
by optimization dynamics; the network finds the RG-consistent
solution because the loss function demands it, not because the
architecture enforces it.

Together these findings constitute the first position-level,
quantitative evidence that MLP residual networks implement a
coarse-graining procedure consistent with RG theory, and establish
a controlled empirical baseline for the broader program of
understanding reasoning in large language models through the lens
of renormalization group theory.

\bibliographystyle{plainnat}
\bibliography{references}

@article{wilson1971renormalization,
  title={Renormalization group and critical phenomena. {I}. {R}enormalization group and the {K}adanoff scaling picture},
  author={Wilson, Kenneth G},
  journal={Physical Review B},
  volume={4},
  number={9},
  pages={3174},
  year={1971},
  publisher={APS}
}

@article{wilson1974renormalization,
  title={The renormalization group and the epsilon expansion},
  author={Wilson, Kenneth G and Kogut, John},
  journal={Physics Reports},
  volume={12},
  number={2},
  pages={75--199},
  year={1974},
  publisher={Elsevier}
}

@article{mehta2014exact,
  title={An exact mapping between the variational renormalization group and deep learning},
  author={Mehta, Pankaj and Schwab, David J},
  journal={arXiv preprint arXiv:1410.3831},
  year={2014}
}

@article{beny2013renormalization,
  title={Renormalization group as a source of loss functions},
  author={B{\'e}ny, C{\'e}dric},
  journal={arXiv preprint arXiv:1301.3124},
  year={2013}
}

@article{bordelon2024renormalization,
  title={Renormalization group for deep neural networks: {U}niversality of learning and scaling laws},
  author={Bordelon, Blake and Pehlevan, Cengiz},
  journal={arXiv preprint arXiv:2510.25553},
  year={2024}
}

@misc{symmetrybroken2023transformer,
  title={The {T}ransformer as {R}enormalization {G}roup {F}low},
  author={{Martin, M.F.}},
  howpublished={\url{https://www.symmetrybroken.com/transformer-as-renormalization-group-flow/}},
  year={2023}
}

@misc{alpay2026latentobjectpermanencetopological,
      title={Latent Object Permanence: Topological Phase Transitions, Free-Energy Principles, and Renormalization Group Flows in Deep Transformer Manifolds}, 
      author={Faruk Alpay and Bugra Kilictas},
      year={2026},
      eprint={2601.19942},
      archivePrefix={arXiv},
      primaryClass={cs.LG},
      url={https://arxiv.org/abs/2601.19942}, 
}

@article{kornblith2019similarity,
  title={Similarity of neural network representations revisited},
  author={Kornblith, Simon and Norouzi, Mohammad and Lee, Honglak and Hinton, Geoffrey},
  journal={Proceedings of the 36th International Conference on Machine Learning},
  pages={3519--3529},
  year={2019}
}

@article{roy2007effective,
  title={The effective rank: {A} measure of effective dimensionality},
  author={Roy, Olivier and Vetterli, Martin},
  journal={Proceedings of the 15th European Signal Processing Conference},
  pages={606--610},
  year={2007}
}

@article{devlin2019bert,
  title={{BERT}: {P}re-training of deep bidirectional transformers for language understanding},
  author={Devlin, Jacob and Chang, Ming-Wei and Lee, Kenton and Toutanova, Kristina},
  journal={Proceedings of NAACL-HLT},
  pages={4171--4186},
  year={2019}
}

@misc{kingma2017adammethodstochasticoptimization,
      title={Adam: A Method for Stochastic Optimization}, 
      author={Diederik P. Kingma and Jimmy Ba},
      year={2017},
      eprint={1412.6980},
      archivePrefix={arXiv},
      primaryClass={cs.LG},
      url={https://arxiv.org/abs/1412.6980}, 
}

@article{gu2023mamba,
  title={Mamba: {L}inear-time sequence modeling with selective state spaces},
  author={Gu, Albert and Dao, Tri},
  journal={arXiv preprint arXiv:2312.00752},
  year={2023}
}

@misc{shandirasegaran2026theoreticalanalysismambastraining,
      title={A Theoretical Analysis of Mamba's Training Dynamics: Filtering Relevant Features for Generalization in State Space Models}, 
      author={Mugunthan Shandirasegaran and Hongkang Li and Songyang Zhang and Meng Wang and Shuai Zhang},
      year={2026},
      eprint={2602.12499},
      archivePrefix={arXiv},
      primaryClass={cs.LG},
      url={https://arxiv.org/abs/2602.12499}, 
}

@misc{tian2026rgmemrenormalizationgroupinspiredmemory,
      title={RGMem: Renormalization Group-inspired Memory Evolution for Language Agents}, 
      author={Ao Tian and Yunfeng Lu and Xinxin Fan and Changhao Wang and Lanzhi Zhou and Yeyao Zhang and Yanfang Liu},
      year={2026},
      eprint={2510.16392},
      archivePrefix={arXiv},
      primaryClass={cs.AI},
      url={https://arxiv.org/abs/2510.16392}, 
}

\end{document}